\documentclass[conference]{IEEEtran}

\usepackage{float}

\usepackage{cite}
\usepackage{amsmath,amssymb,amsfonts}
\usepackage{algorithmic}
\usepackage{graphicx}
\usepackage{multirow}
\usepackage{subcaption}
\usepackage{booktabs}
\usepackage{textcomp}
\usepackage{xcolor}

\usepackage{physics}
\usepackage{amsmath}
\usepackage{tikz}
\usepackage{mathdots}
\usepackage{yhmath}
\usepackage{cancel}
\usepackage{color}
\usepackage{siunitx}
\usepackage{array}
\usepackage{multirow}
\usepackage{amssymb}
\usepackage{gensymb}
\usepackage{tabularx}
\usepackage{extarrows}
\usepackage{booktabs}
\usetikzlibrary{fadings}
\usetikzlibrary{patterns}
\usetikzlibrary{shadows.blur}
\usetikzlibrary{shapes}

\def\BibTeX{{\rm B\kern-.05em{\sc i\kern-.025em b}\kern-.08em
    T\kern-.1667em\lower.7ex\hbox{E}\kern-.125emX}}
\onecolumn
\begin{document}
\title{A Data-driven Recommendation Framework for Optimal Walker Designs \\ 
}
\author{\IEEEauthorblockN{Advaith Narayanan}
\IEEEauthorblockA{\textit{Leigh High School} \\
\textit{advaith.narayanan20@gmail.com}}
}

\maketitle
\begin{abstract}
The rapidly advancing fields of statistical modeling and machine learning have significantly enhanced data-driven design and optimization. This paper focuses on leveraging these design algorithms to optimize a medical walker, an integral part of gait rehabilitation and physiological therapy of the lower extremities. To achieve the desirable qualities of a walker, we train a predictive machine-learning model to identify trade-offs between performance objectives, thus enabling the use of efficient optimization algorithms. To do this, we use an Automated Machine Learning model utilizing a stacked-ensemble approach shown to outperform traditional ML models. However, training a predictive model requires vast amounts of data for accuracy. Due to limited publicly available walker designs, this paper presents a dataset of more than 5,000 parametric walker designs with performance values to assess mass, structural integrity, and stability. These performance values include displacement vectors for the given load case, stress coefficients, mass, and other physical properties. We also introduce a novel method of systematically calculating the stability index of a walker. We use MultiObjective Counterfactuals for Design (MCD), a novel genetic-based optimization algorithm, to explore the diverse 16-dimensional design space and search for high-performing designs based on numerous objectives. This paper presents potential walker designs that demonstrate up to a 30\% mass reduction while increasing structural stability and integrity. This work takes a step toward the improved development of assistive mobility devices.
\end{abstract}
\section{Introduction}
A walker is an assistive device for ambulation, typically used by individuals with a loss of internal balance or an impairment in the lower extremities. More than 4 million people in the United States alone use a walker every day~\cite{gell2015mobility}. Walkers enhance independent mobility, promote physical activity, elevate emotional well-being, and reduce falls and hospital visits~\cite{meng2019use}. However, more than 41,000 walker-related injuries occur every year.\cite{stevens2009Unintentional}. Therefore, taking steps towards optimizing a walker may have a significant impact on public health and life expectancy, enabling people to work longer and economically prosper on both a micro and macro scale\cite{NBERw8808}.

The main purpose of this paper is to create a tool that recommends medical walker design modifications, i.e. an optimizer. To optimize a walker, we use data-driven methods to analyze and develop high-performing designs. Traditional methods such as physical trial-and-error testing are very inefficient and expensive. Using data-driven methods, surrogate models can be trained to help designers and researchers optimize walkers without the need for extensive physical experimentation. Surrogate models can efficiently predict design performance and highlight data trends that are valuable for optimization. We aim to not only create the data needed for such surrogate model training but also utilize the capabilities of novel optimization algorithms and such data-driven approaches to optimize walkers. 

\begin{figure}[h]
    \centering
    \includegraphics[scale=.58]{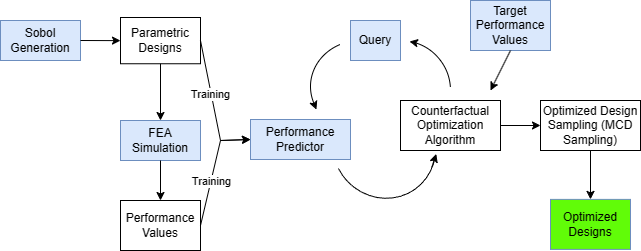}
    \caption{Illustration of the dataset generation and optimization process}
    \label{fig:blockDiagram}
\end{figure}

An optimizer needs to understand how design modifications can affect performance values. Hence, we train a surrogate model (performance predictor) in Section \ref{optimization}, where the optimizer can “query” the predictor to understand the trends in the performance values. For a surrogate model to be accurate, vast amounts of data are needed. So, we generate a dataset using a parametric design and simulations in Section \ref{datasetGeneration}. This approach is outlined in Figure \ref{fig:blockDiagram}.

Key contributions of this paper are listed below:
\begin{itemize}
    \item We create a dataset of 5006 parametric walker designs. Each design contains 16 parameters along with 8 performance values simulated under a single load case representative of multiple use cases.
    \item We train a surrogate model on the dataset using Automated Machine Learning frameworks. The model predicts the 8 performance values of the 2-wheeled walker.
    \item We introduce a method of systematically calculating both the static and “dynamic” stability of a walker based on its mass and dimensions. 
    \item We use the surrogate model to optimize walker parameters using a novel counterfactual search algorithm: Multiobjective Counterfactuals for Design. We address three key objectives during the optimization: Weight, Structural integrity, and Stability. 
    \item We provide an open-source GitHub repository that includes all the code, data, and models: github.com/AdvaithN1/Walker-Optimization.git
\end{itemize}

\section{Background} \label{background}
This paper investigates the use of data-driven optimization for assistive mobility walkers,  specifically the 2-wheeled walker. This section discusses related prior works and an overview of key concepts.

\subsection{Walker Structure Optimization}
Walkers that have minimal weight, maximum stability, and maximum durability are generally desirable. However, excessive weight reductions can result in lower stability and durability. We strive for the optimal compromise between these conflicting objectives.

Since the 1950s\cite{walkerPat}, many different modern walker designs emerged to fit the needs of their users. Rollator walkers (4-wheeled walkers) are typically easier to move but less stable. “Fixed-frame” walkers (walkers with no wheels) are more stable but are harder to move due to the need to lift the entire walker during ambulation. The 2-wheeled walker lies in between, with a balance between stability and mobility. 

Many studies have attempted to broaden the use of various walkers to a larger target group. For example, Mostofa \textit{et al.}\cite{mostofa2021smart} attempt to create and optimize a walker for those with both visual and mobility impairments by studying 4 different configurations. Others have attempted to create “smart” walkers. Shin \textit{et al.}\cite{shin2016smartwalker} present a walker with autonomous assistive capabilities. However, to our knowledge, there is little data-driven research on the optimization of the fundamental aspects of a walker, hence the creation of this paper. To gauge the performance of a walker, we leverage the efficiency and accuracy of Computer-Aided Design (CAD). The exponential growth of computational power over the last few decades allows for relatively accurate estimates of the stresses and deflections walker designs undergo. We use Finite Element Analysis (FEA), a numerical simulation technique, to calculate these values.

\subsection{Machine Learning/AutoML/AutoGluon}
Machine Learning (ML) is a subfield of Artificial Intelligence (AI) that is used to learn patterns in data. This paper uses ML for regression tasks to identify patterns in simulation data. Automated Machine Learning (AutoML) automates and streamlines many of these tasks by eliminating the need for manual hyperparameter optimization, data preprocessing, etc\cite{feurer2015efficient}. AutoGluon is an AutoML framework that uses a stacked-ensemble approach, meaning that it trains various types of models on the same data and “combines” them into one ensemble model. Regenwetter \textit{et al.}\cite{regenwetter2023framed} show that AutoGluon outperforms traditional hyperparameter-tuned ML models in certain parametric design tasks, especially in bike frame performance prediction. Due to the many similarities between walker frame structures and bicycle frames, AutoGluon is anticipated to be a similarly performing regression model, hence our choice of AutoGluon.

\subsection{Multiobjective Counterfactuals for Design}
A counterfactual is a hypothetical statement or question that leads to a change in the outcome. However, this paper is concerned with the inverse counterfactual: What would hypothetically need to happen to lead to a specific outcome? “Counterfactual Search” finds the conditions (parameter changes) that lead to a pre-defined outcome. Only data-driven design optimization allows for such precise outcome targets.

Regenwetter \textit{et al.}\cite{regenwetter2023counterfactuals} propose a counterfactual search approach using Multiobjective Counterfactuals for Design (MCD), a novel genetic-based optimization algorithm. Genetic algorithms are especially useful in design optimization due to their compatibility with multiple objectives and their Pareto-optimality property. Pareto-optimality ensures that the output designs are not “worse” than one another—no design dominates over another, thus increasing the value of the optimization results. MCD includes various hyperparameters catered towards design optimization, discussed in section \ref{optHyp}. Most importantly, unlike general optimization algorithms such as gradient descent, MCD uses iterative and targeted optimization, which enables the use of strict constraints on the performance of the design. 
\section{Dataset Generation} \label{datasetGeneration}
This section discusses the methodology behind the dataset generation, including parameterization, design validation, loads, and material.

\subsection{Walker Parameterization and Modeling}
To procedurally generate a walker dataset, we create a parametric representation that balances expressivity and reliability. To create a parametric representation that achieves this balance, we make key simplifications to the 2-wheeled walker model:
\begin{enumerate}
    \item We assume that the wheels act as a roller joint with minimal rolling friction. We do not model a wheel but rather implement the wheels’ functionality invisibly into the front legs.
    \item We assume that, aside from the front crossbeams, the frame is laterally symmetric excluding the wheels.
    \item We do not consider the folding or height-adjusting mechanisms of the walker frame.
    \item We do not consider tube warping and screw joints at tube junctions.
    \item We assume that the overall base of the walker is a rectangle.
\end{enumerate}
\begin{figure}
    \centering
    \includegraphics[scale=.3]{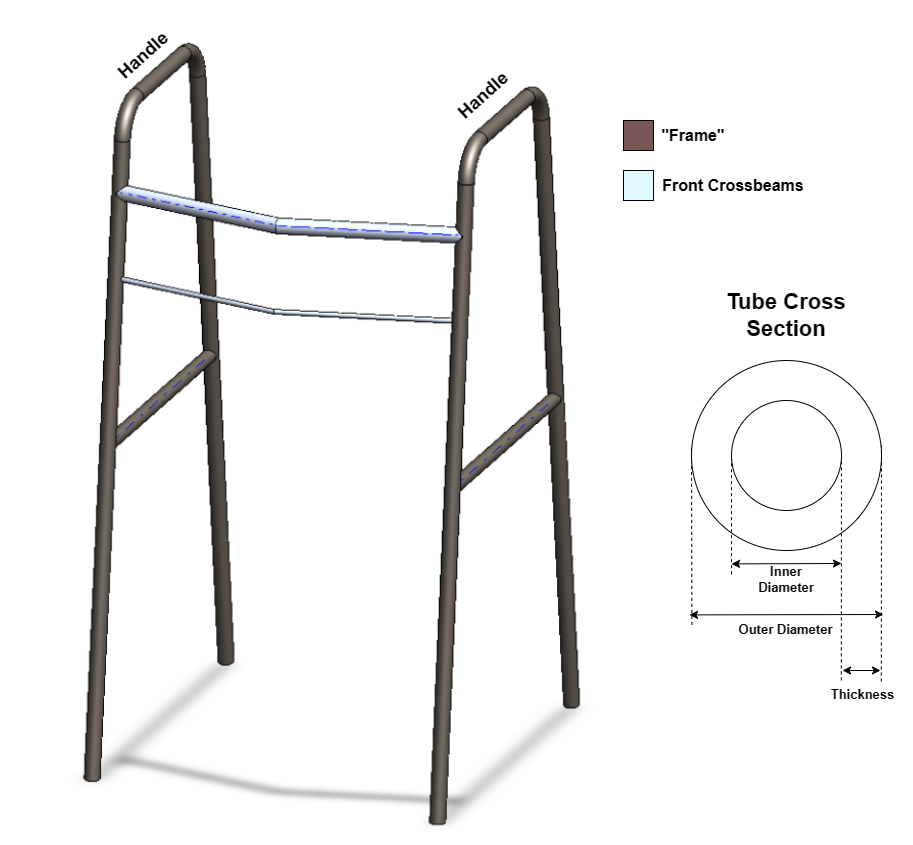}
    \caption{Illustration of Walker Terminology}
    \label{fig:walkerParts}
\end{figure}
These simplifications allow us to completely and reliably define a walker with 16 continuous parameters without a significant compromise on the expressivity of the representation. These parameters include the overall dimensions of the frame, tube thicknesses, and tube diameters as summarized in Table \ref{tab:params}. Additionally, we include 2 categorical parameters to serve as material identifiers for different parts of the walker. The material parameters are discussed in more detail in section \ref{matSel}. \begin{table}
\centering  
\begin{tabular}{|l|l|l|l|l|}  
\hline  
Parameter Type      & Data Type   & Count\\
\hline  
Geometry Relations  & Continuous  & 9    \\
Tube Inner Diameter & Continuous  & 3     \\
Tube Thickness      & Continuous  & 2    \\
Material            & Categorical & 2    \\
\hline 
Total               &             & 16 \\
\hline  
\end{tabular}
\caption{Categorical summary of the parametric model design space.}
\label{tab:params}
\end{table}

To promote diversity in the dataset, we choose parameter ranges that encompass a broad space of reasonable designs without compromising on data point density. We use a Sobol sequence, a quasi-random sampling algorithm, to efficiently and uniformly sample the multidimensional design space. This is described in more detail in the Appendix. 

\begin{figure}
    \centering
    \includegraphics[scale=.28]{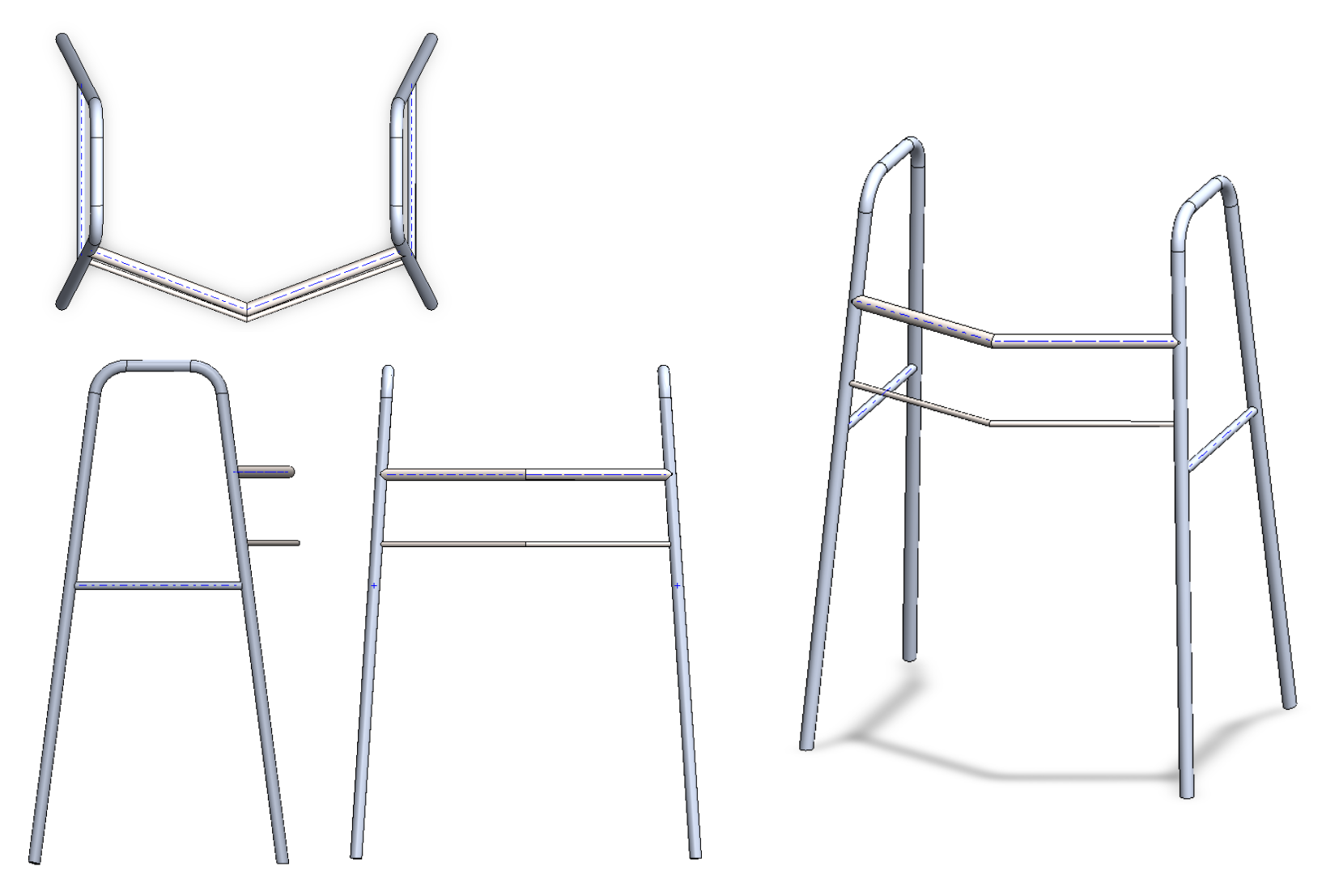}
    \caption{Multiple views of example walker}
    \label{fig:walkerViews}
\end{figure}

We create an adaptive 3D parametric model using SolidWorks and analyze it in Section \ref{paramModelAnalysis}. Figure \ref{fig:walkerViews} illustrates this 3D model. The template includes all of the parameters discussed above. We use the Design Study feature in SolidWorks to import data and generate model variations.

\subsection{Geometric Feasibility}
Following the sampling process, we drop designs with potential geometric infeasibilities. These checks for invalid designs include the following:
\begin{itemize}
    \item Length parameters must be positive.
    \item Crosspiece positions must not exceed the walker height.
    \item Junctions with varying outer tube diameters must follow checks for improper connections.
    \item The outer diameter must not exceed a certain threshold.
\end{itemize}
Approximately 66\% of generated designs were systematically dropped due to one or more unsatisfied feasibility checks. Even with these checks in place, 394 designs failed to build during simulation runtime due to miscellaneous geometry errors, which resulted in a total of 5006 valid models. 

\subsection{Parametric Model Analysis} \label{paramModelAnalysis}
\begin{figure}
    \centering
    \includegraphics[scale=0.3]{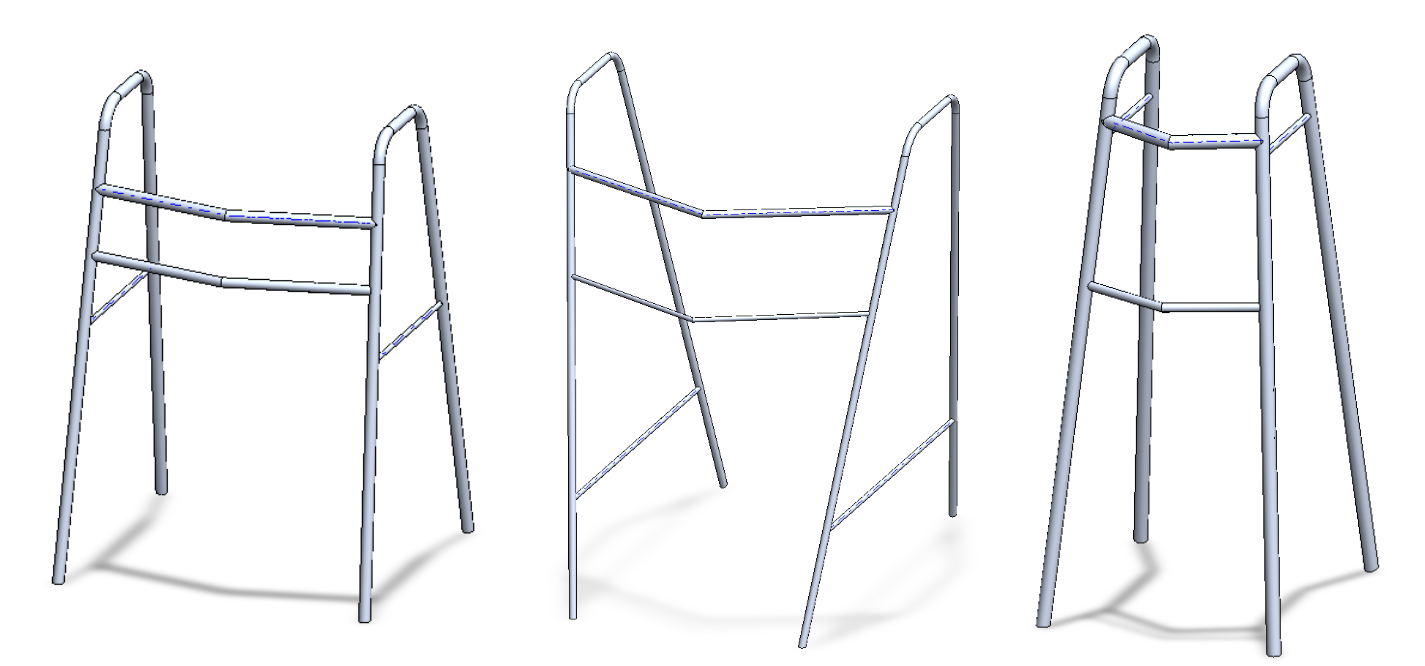}
    \caption*{Example 1\ \ \ \ \ \ \ \ \ \ \ \ \ \ \ \ \ \ \ \ \ \ \ \ \ \ \ \ Example 2\ \ \ \ \ \ \ \ \ \ \ \ \ \ \ \ \ \ \ \ \ \ \ \ \ \ \ \ Example 3}
    \label{fig:3walkersExs}
\end{figure}
We examine the capabilities of the parametric model through parameter manipulation. We analyze 3 different configurations.

\subsubsection*{Example Walker 1}
This design resembles the typical walker, with an approximately 170\degree\ bend in the front crossbeams and a slight tilt in the side walker frames. Being a relatively accurate recreation of the publicly available walker, this design validates the capabilities of the parametric model to construct a realistic walker. We will refer to this design as the ‘original’ walker.
\subsubsection*{Example Walker 2}
This is a rather odd-looking walker, with particularly thin tubes and a protruding frame. Its peculiarities, however, indicate the diverse possibilities of this parametric model. Under a downward load, we see that the front crossbeams would undergo significant amounts of tension, thereby reducing the shear stress on the rest of the frame. However, this design is likely unfeasible due to its extremely thin frame and stability issues. 
\subsubsection*{Example Walker 3}
This example highlights the extremes of the parameters. With exceptionally thick tubes, this walker resembles a pyramid-like structure. This walker would be highly durable and stable due to its shape, but likely unfeasible due to the narrow space between the handles, suggestive of the necessary strict constraints that need to be enforced during parameter manipulation.

\subsection{Material Selection} \label{matSel}
The parametric walker model contains 2 categorical parameters. The first represents the material of the front two cross pieces, and the other represents the rest of the walker (outer frame). Walkers are typically made from aluminum due to its lightweight properties. However, we experiment with the use of steel and titanium, both of which have a higher yield and tensile strength with the tradeoff of higher mass density, as shown in Table \ref{tab:materialInfo}. For each material, we use standard industry alloys, such as the 6061-T6 Aluminum Alloy, AISI 4130 Steel Alloy, and the Ti-6Al-4V Titanium Alloy.

\begin{table}[h]
\renewcommand{\arraystretch}{1.5}
\centering
\begin{tabular}{|l|l|l|l|}
\hline
Material               & Aluminum & Steel & Titanium \\
\hline
Elastic Modulus (GPa)  & 69       & 205   & 105      \\
\hline
Poisson’s Ratio        & 0.330    & 0.285 & 0.310    \\
\hline
Shear Modulus (GPa)    & 26       & 80    & 41       \\
\hline
Density (kg/m3)        & 2700     & 7850  & 4429     \\
\hline
Tensile Strength (MPa) & 310      & 731   & 1050     \\
\hline
Yield Strength (MPa)   & 275      & 460   & 827     \\
\hline
\end{tabular}
\caption{Material Properties}
\label{tab:materialInfo}
\end{table}

\subsection{Simulation Setup and Loads}
\begin{figure}
\centering
\begin{subfigure}{.5\textwidth}
  \centering
  \includegraphics[width=1\linewidth]{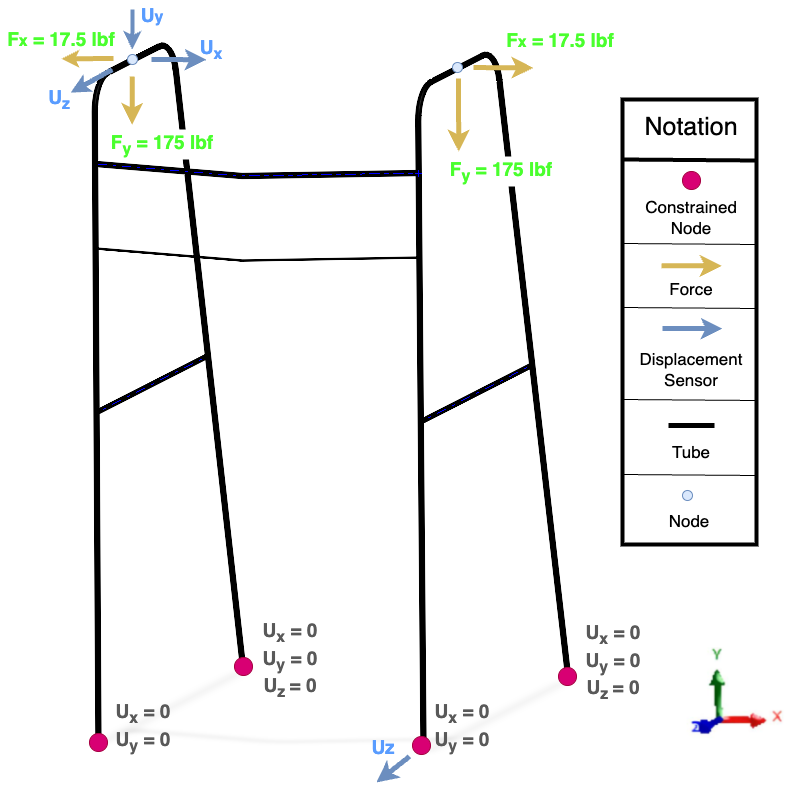}
  \caption{Illustration of the load case, node constraints, and sensors.}
  \label{fig:labeledWalker}
\end{subfigure}%
\begin{subfigure}{.5\textwidth}
  \centering
  \includegraphics[width=0.8\linewidth]{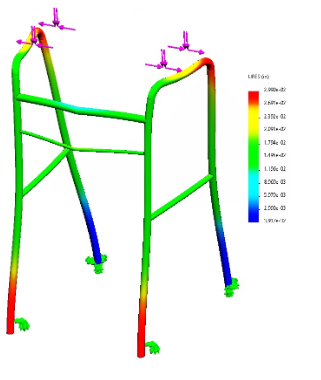}
  \caption{Example walker with amplified deformation}
  \label{fig:displacedWalker}
\end{subfigure}
\label{fig:test}
\caption{}
\end{figure}

To properly assess the structural characteristics of a walker under a broad range of use cases, we choose a single load case as a representation of both normal and eccentric loading due to the available computational budget. We use the publicly listed maximum weight of 350 lbf as the downwards force on the handles and a 17.5 lbf lateral outwards force on the handles to represent eccentric loading, as illustrated in Figure \ref{fig:labeledWalker}. 



We run one simulation for every input design. For each simulation, the deformed model is captured. Beam deflections are often very small and visually imperceptible, so Figure \ref{fig:displacedWalker} displays different views of an example walker with overridden amplified deformation values. The handle and front wheels experience the greatest deformation. So, using the sensors shown in Figure \ref{fig:labeledWalker}, we measure the x, y, and z displacements of the handles and the displacement of the front wheels. We measure the total mass and the position of the center of mass relative to the midpoint of the rectangular base of the walker. Finally, we measure the minimum safety factor, for a total of 8 performance values.

To run the simulation, we use Finite Element Analysis (FEA), a method of calculating physical model behavior under defined loads through the analysis of interactions between smaller components. This is done by dividing the model into small 3D polygon elements, creating a mesh. Separate physical calculations are performed on each element, and the process is repeated for multiple steps. Smaller element sizes create a “finer” mesh, which more accurately represents the real object, but is computationally slower. We select a mesh resolution of 0.05 inches as the minimum element size and 1 inch as the maximum element size. These values were selected to maintain relative accuracy while keeping within the strict computational budget. We use SolidWork’s “Blended Curvature-Based Mesh,” which reduces the element size in sections of high curvature. Although higher-order meshing could be used to refine the mesh representation without increasing computational complexity, compatibility and feature availability issues arise, thus making linear meshing the only feasible option.

\subsection{Performance Space Visualization}

\begin{figure}
    \centering
    \includegraphics[scale=0.21]{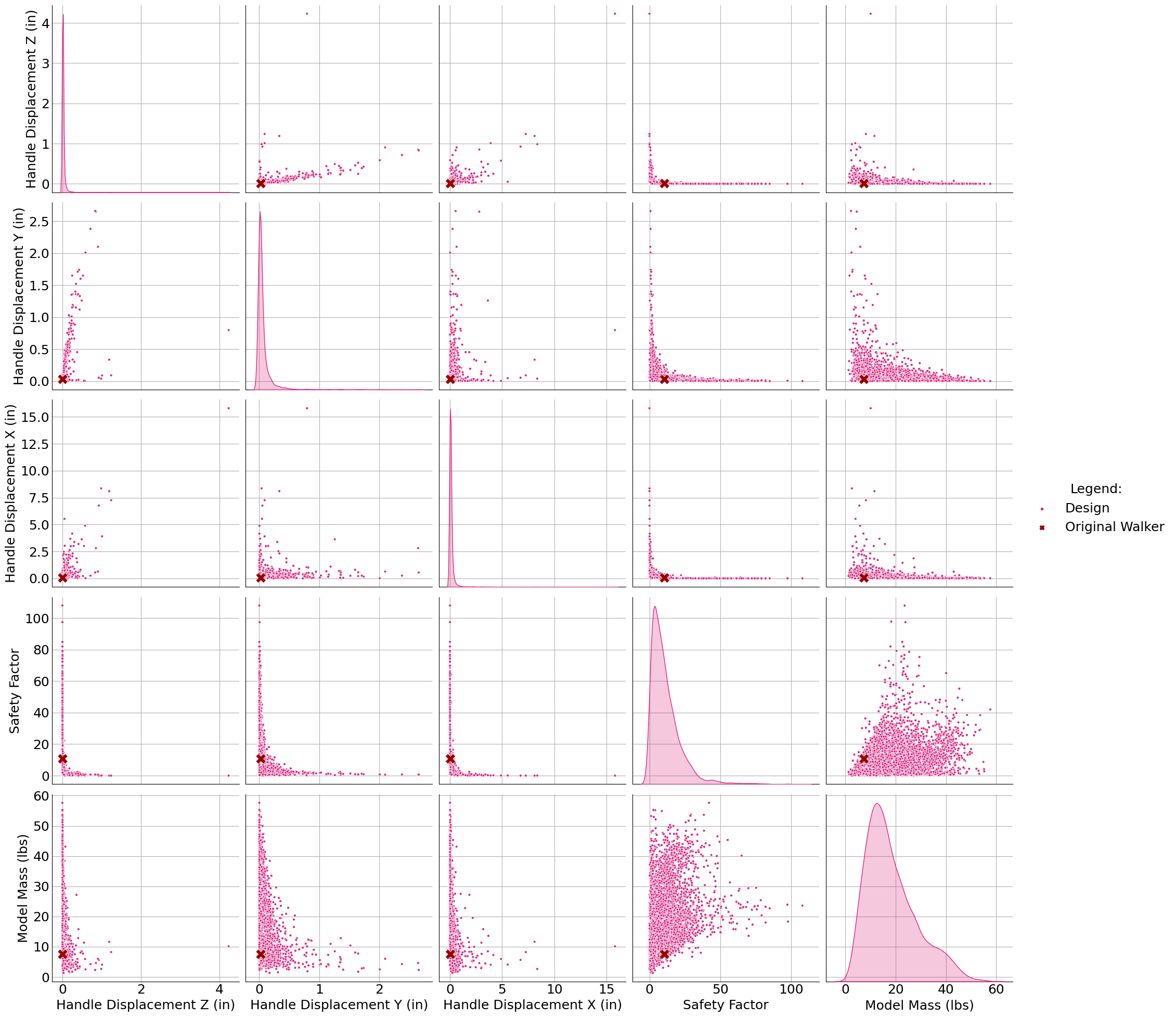}
    \caption{Performance Space Visualization with the original design and Kernel Density Estimate Plots}
    \label{fig:perfSpaceVisual}
\end{figure}

In data-driven design, there is an important distinction between the design space and performance space. The design space is a sort of multidimensional parameterization space, where each point defines a specific walker. The performance space is a similar multidimensional space that includes the performance values derived from the defining characteristics (i.e. parameters) of each walker. To intuitively validate the trends in the performance values, we visualize the performance space in Figure \ref{fig:perfSpaceVisual} using 5 out of the 8 performance values collected. Before plotting, we take the absolute value of the deflection values. Each plot can be thought of as a 2-dimensional “Cross Section” of the 8-dimensional space, with each point representing the performance values of one design in the dataset. Across the diagonal, we show the Kernel Density Estimation (KDE) plot, which estimates the frequency of the values for a certain performance value in the dataset. Finally, we represent the original walker design (the publicly listed walker design) on the plots for comparison.

Upon analysis of the performance space, the strong inverse correlation between the Model Mass and the Handle Displacements is found. This validates the intuition that designs with greater mass tend to deflect less easily. Additionally, there is a clear relationship between the Handle Displacements and the Safety Factor (SF): unsurprisingly, as the SF increases, the displacements decrease. However, quite interestingly, there is not much correlation between Model Mass and SF, likely due to the numerous potential configurations that significantly reduce the load-bearing capability and overcome the effects of mass.

Upon analysis of the KDE plots, we find that the mass of the walkers in the dataset is right-skewed with a peak at around 12 lbs. The safety factor is a similar shape with a peak at around 6. This is to be expected; the thickness values approximately linearly distributed, but the mass and the safety factor are more proportionate to the square of the thickness.

\section{Optimization} \label{optimization}
This section discusses the methodology behind the optimization of the 2-wheeled walker using the dataset. 

\subsection{Surrogate Model}

\begin{figure}
     \centering
     \includegraphics[scale=0.63]{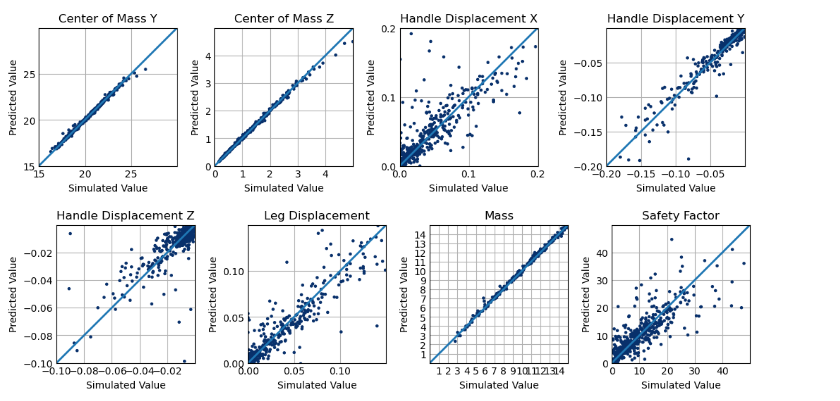}
     \caption{Visualization of the accuracy of AutoGluon Predictions}
     \label{fig:predAccuracy}
 \end{figure}

Optimizing and finding an optimized walker typically involves more than 100,000 simulation queries. Running this many simulations is impractical, especially due to the large design space. So, we leverage an AutoGluon regression model that “learns” from a smaller set of simulations and generalizes it to a continuous function for each performance value. To validate the accuracy of the regressor model, we separate a test dataset not seen by the regressor during training. We compare the predicted values to the actual simulated values, as shown in Figure \ref{fig:predAccuracy}. 

Handle X, Y, and Z are handle deflections (displacements) for their respective dimensions. Center Y and Center Z are the coordinates of the position of the center of mass with respect to the center of the base of the walker. 

Upon visual analysis, we find that the mass and center of mass position values are predicted at exceptional accuracy. However, visual analysis is insufficient for precisely gauging the accuracy. The coefficient of determination value (R-squared value) is a quantitative measurement of how well a regressor can predict a certain value, with the highest/best value being 1. A negative R-squared value indicates that a linear regressor can perform better than the predictive model. The R-squared values are listed in Table \ref{tab:rsqaureds}.

\begin{table}[]
\renewcommand{\arraystretch}{1.5}
\centering
\begin{tabular}{|c|c|c|c|c|c|c|c|c|c|}
\hline
\centering
Predicted Value & Mass & Handle X Displ & Handle Y Displ & Handle Z Displ & Safety Factor & Center Y & Center Z & Leg Displ \\
\hline
R-Squared Value & 0.9989 & -4.9753        & 0.8249                             & 0.2647         & 0.6962        & 0.9937   & 0.9934   & 0.8220   \\
\hline
\end{tabular}
\caption{Coefficient of Determination values for each performance value}
\label{tab:rsqaureds}
\end{table}

Most R-squared values are above 0.5, which is an acceptable accuracy for our work. However, the Handle X and Z values are far below 0.5. So, we minimally use these performance values when querying the predictor.

\subsection{Performance Objectives}
This paper addresses 3 main objectives during the optimization of the 2-wheeled walker: weight, structural integrity, and stability. 

During walker-assisted ambulation, the user must lift or, in the 2-wheeled walker case, must exert a torque to lift the rear legs before rolling the walker forward. This torque or force is undesirably counteracted by the gravitational force on the walker. Hence, mass is a key objective to be minimized to ensure minimal physical strain induced on the user. We directly measure the mass of each design as a part of the performance values and aim to achieve a weight of the publicly listed 7.5 lbs or less.

Structural integrity is a measure of how well a system is resistant to breaking or deflecting. Maximizing structural integrity is an integral objective for the safety of the user and ergonomics—the consequences of small deflections are amplified through prolonged use. Many performance values correlate to this objective. For example, the Minimum Safety Factor (SF) is a measure of how well a design can handle a load case. A minimum SF of 1 indicates that the structure will fail exactly when it reaches the pre-defined loads, while greater than 1 indicates that the structure exceeds the necessary properties required to keep from failing at the load case. Thus, we attempt to maximize the safety factor during optimization. Displacement values are the deflection values at certain points on the walker. Due to their inverse correlation to structural integrity, we aim to minimize these values.

\begin{figure}
    \centering
    \includegraphics[scale=0.5]{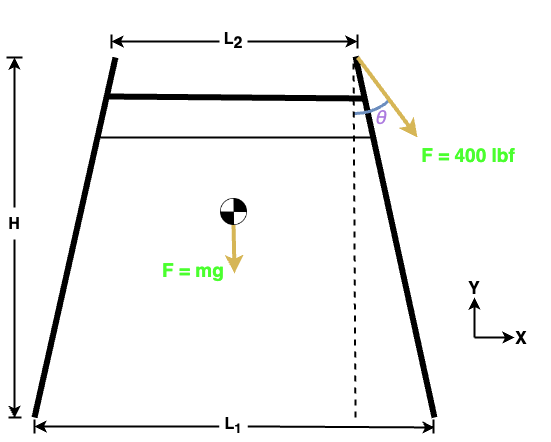}
    \caption{Front-view illustration of the load case used to calculate stability. Frame angles are exaggerated for visual clarity.}
    \label{fig:stabilityLoad}
\end{figure}

Lastly, we place stability as one of the key objectives. However, none of the performance values measured from the simulations directly correlate to stability. To assess stability, we exert a hypothetical force outwards and on the walker handles to find the angle to the vertical at which the walker will start to tip, as shown in Figure \ref{fig:stabilityLoad}. This setup is chosen due to its realistic representation of uneven lateral forces exerted on the walker, typically present during leaning or in extreme cases of erratic ambulation. Designs with a greater tipping angle are generally more stable. We balance the forces and torques at equilibrium to derive an equation for the tipping angle from available parameters:
\begin{gather}
\phi =\arcsin\left(\frac{mgL_{1}}{2F\sqrt{H^{2} +\left(\frac{L_{1} -L_{2}}{2}\right)^{2}}}\right)\\
\theta =\phi +\arctan\left(\frac{L_{1} -L_{2}}{2H}\right)
\end{gather}
where m is the mass, g is the gravitational acceleration, L1 is the lateral distance between the legs, L2 is the distance between handles, H is the vertical height, F is the force, and $\theta$ is the tipping angle. Analysis of the equation reveals that when $\phi$ is out of the domain (i.e. undefined), there is no theta such that the walker will tip. If the force F is minimal, then some walkers may not have a defined tipping angle, as no angle will tip the walker. If the force is very large, then variations in the design will cause little to no variation in the tipping angle. So, we choose the force F to be 400 lbf to ensure sufficient tipping angle deviation for accurate optimization, but such that the tipping angle is defined for almost all designs. 

However, the tipping angle is a measure of static stability. Once the walker starts tipping, the above analysis is irrelevant. We define “dynamic” stability, in the context of a walker, to be a measure of how well the walker returns to the equilibrium position after a tipping disturbance. Analysis of the forces reveals that a lower center of mass exerts a greater restoring torque as the walker is tipping. Therefore, to achieve maximum “dynamic” stability, the elevation of the center of mass of the walker must be minimized. 

\subsection{Design Exploration Constraints}

In the design space, invalid, inaccurate, and infeasible designs are ubiquitous, so accordingly strict constraints must be enforced on the optimizer.

Since the surrogate model (predictor) is less reliable when queried designs far from its training data, we enforce constraints on the optimizer and validate optimized designs through direct simulation. This way, inaccuracies can be minimized.

Additionally, constraints must be set in place to ensure structure practicality. For example, the distance between handles must not significantly vary due to the precise nature of walker ergonomics. Changes in this distance may result in shoulder blade discomfort and other adverse effects. So, we set the range of the distance between the handles to be significantly stricter than the range constraints for other parameters. 

\subsection{Optimization Hyperparameters} \label{optHyp}
As per typical genetic algorithms, hyperparameters such as Population Size and Generations must be manually set, so we use trial and error to determine the ideal values for these parameters.

MCD’s final sampling process chooses the highest value designs from the generated counterfactuals. For example, if 100 valid counterfactual designs are found (i.e. designs that fit the performance value requirements), the sampling process creates a subset of the highest-value designs from the original 100. The sampling process includes many hyperparameters that we again manually select through trial and error. 

The “original” design is the parametric representation of the typical publicly available walker design, with a weight of 7.5 lbs. This original design will be used in the MCD Case Analyses.

\section{MCD Analysis and Case Studies}
This section examines the various use cases of the optimizer and analyzes the MCD optimization results.

\subsection{Custom and Generic Optimization}
In this paper, we will focus on two fundamental ways of optimization: Generic Optimization, which we define as finding a single design that fits the needs of a broad set of users, and customization, which aims at finding a design that fits the specific needs of an individual. In generic optimization, we demonstrate that the framework can simultaneously optimize for many objectives at once, yielding a highly effective design suitable for a variety of use cases. In customization, we demonstrate the potential for  

In the case of generic optimization, we use a target height and distance between the handles. We set the target height range to be from 32-38 in. We also set the distance between handles to be 19±0.5 in, as per standard walker dimensions. The remaining 14 parameters are less constrained to allow for optimization. Regarding the performance values, we set the target mass value range to less than 6 lbs as opposed to the 7.5 lbs of the original design. We also constrain the deflection, SF, and theta performance values to be “better” than the original design. This way, we generate designs that increase the structural integrity and stability of a walker while decreasing its mass. We generate and demonstrate generically optimized designs in Section \ref{genOptimizedStudies}.

In the case of customization, a user may have specific needs that require further constraints for the optimizer. For example, a target height and handle distance can be specified in line with the user’s needs. Additionally, a user-specific objective can be prioritized over others, adding to the customizability. For example, a user may need additional stability but may not significantly benefit from reduced walker weight. In this scenario, stricter performance value constraints will be placed on those pertaining to stability, i.e. theta, while the rest of the performance values will be less constrained. Using this approach, we illustrate and analyze hypothetical individualized designs in Section \ref{custOptimizedStudies}.

In both optimization scenarios, we do not enforce any constraints on the center of mass positions. These values are only used for physical insight, though they may be helpful for future studies concerning complex stability analysis. We also do not constrain the X and Z Handle displacements due to their low r-squared scores in the surrogate model.

\subsection{Generic Optimization Case Studies} \label{genOptimizedStudies}

\begin{figure}
    \centering
    \includegraphics[scale=0.25]{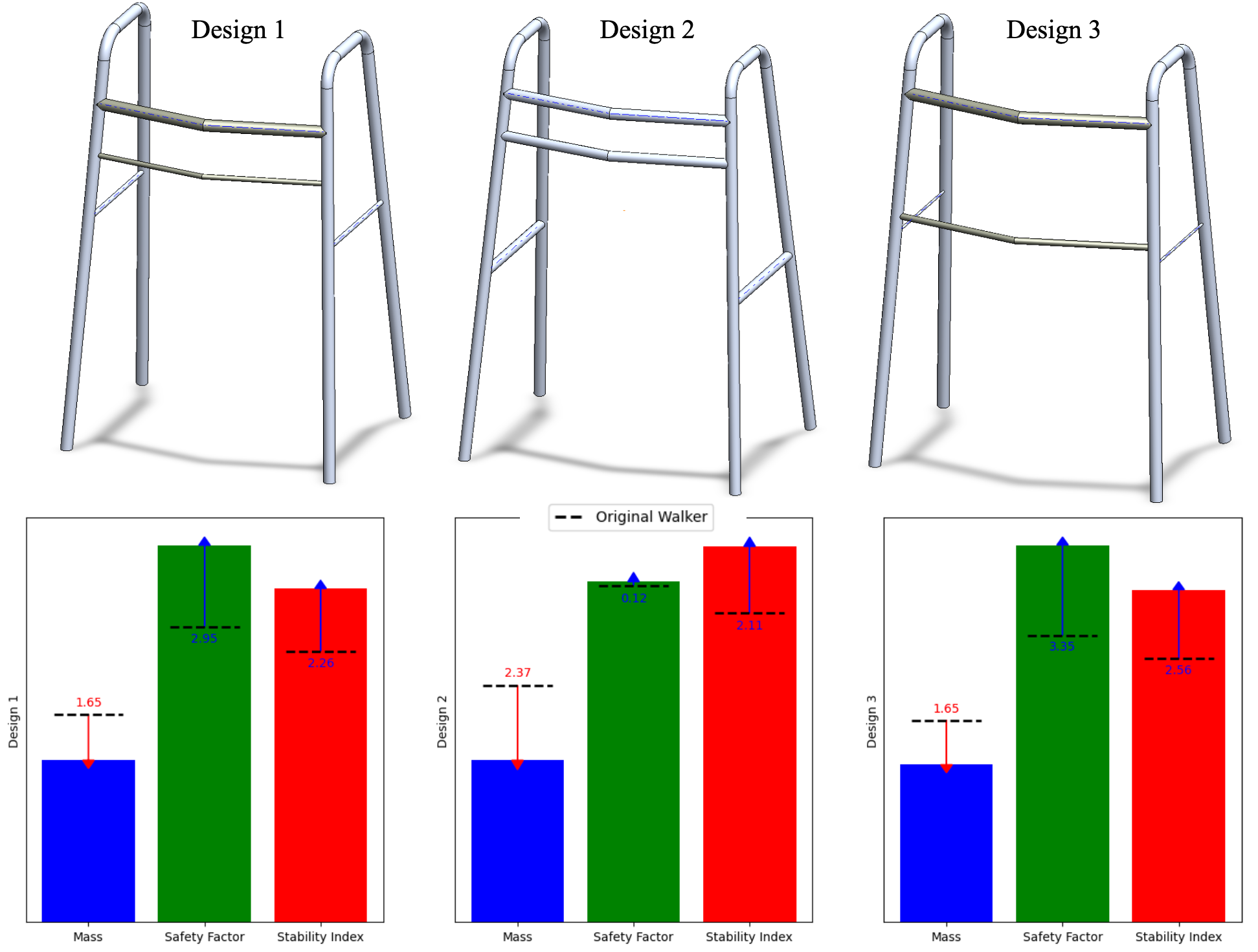}
    \caption{}
    \label{fig:genOptimizedDesigns}
\end{figure}

Please refer to Figure \ref{fig:genOptimizedDesigns} for the following case study analyses.
\\
\subsubsection*{Design 1}

This design’s safety factor, magnitudes of deflection, and theta (stability angle) value are all better than the original design. While most generated counterfactual designs are completely made with aluminum, this design has titanium front crossbeams. Titanium is more dense than aluminum but has more than 3 times the tensile strength. This compensates for the added weight, especially due to high tensile forces acting on the front cross beams during walker-assisted ambulation. However, this design may be costly due to its significant use of titanium. 
\\
\subsubsection*{Design 2}

This design is an outstanding 5.13 lbs, with more than a 30\% decrease in mass from the original 7.5 lbs. However, as shown in Figure \ref{fig:genOptimizedDesigns}, its safety factor is not significantly better than the original design. This design’s lower front crossbeam is unusually thick, with a diameter of 0.755 in. While this does reduce lateral stress, the compromise on the thickness of the frame results in greater deflections in the vertical and longitudinal axes. So, while still feasible, this model compromises structural integrity for a greater decrease in mass. Nevertheless, this design has its practical applications—for instance, it can reduce fatigue under long periods of use under the condition that it does not undergo excessive amounts of force. 
\\
\subsubsection*{Design 3}

This design is 5.85 lbs, but maintains significant increases in safety factor and decreases in the deflection values of up to a factor of 5. Its theta/stability value is greater than any of the designs discussed previously. Quite notably, its front cross beams are again made of titanium, while the rest is aluminum. This trend during the optimization suggests the significant stress that the front crossbeams undergo. Additionally, as visible in the figure, the side crossbeams are extremely thin at a diameter of just 0.2 inches, just above the threshold of the tube validity check. This suggests that the optimizer may have tended towards entirely removing these beams in this specific design. Such observations call for further studies on the practicality of these beams, and their impact on the overall structural integrity. 
\\

We choose design 3 as the most optimal for the generic optimization scenario. This is because of its outstanding safety factor and stability index, yet still maintains a low mass. 

\subsection{Custom Optimization Case Studies} \label{custOptimizedStudies}
\begin{figure}
    \centering
    \includegraphics[scale=0.29]{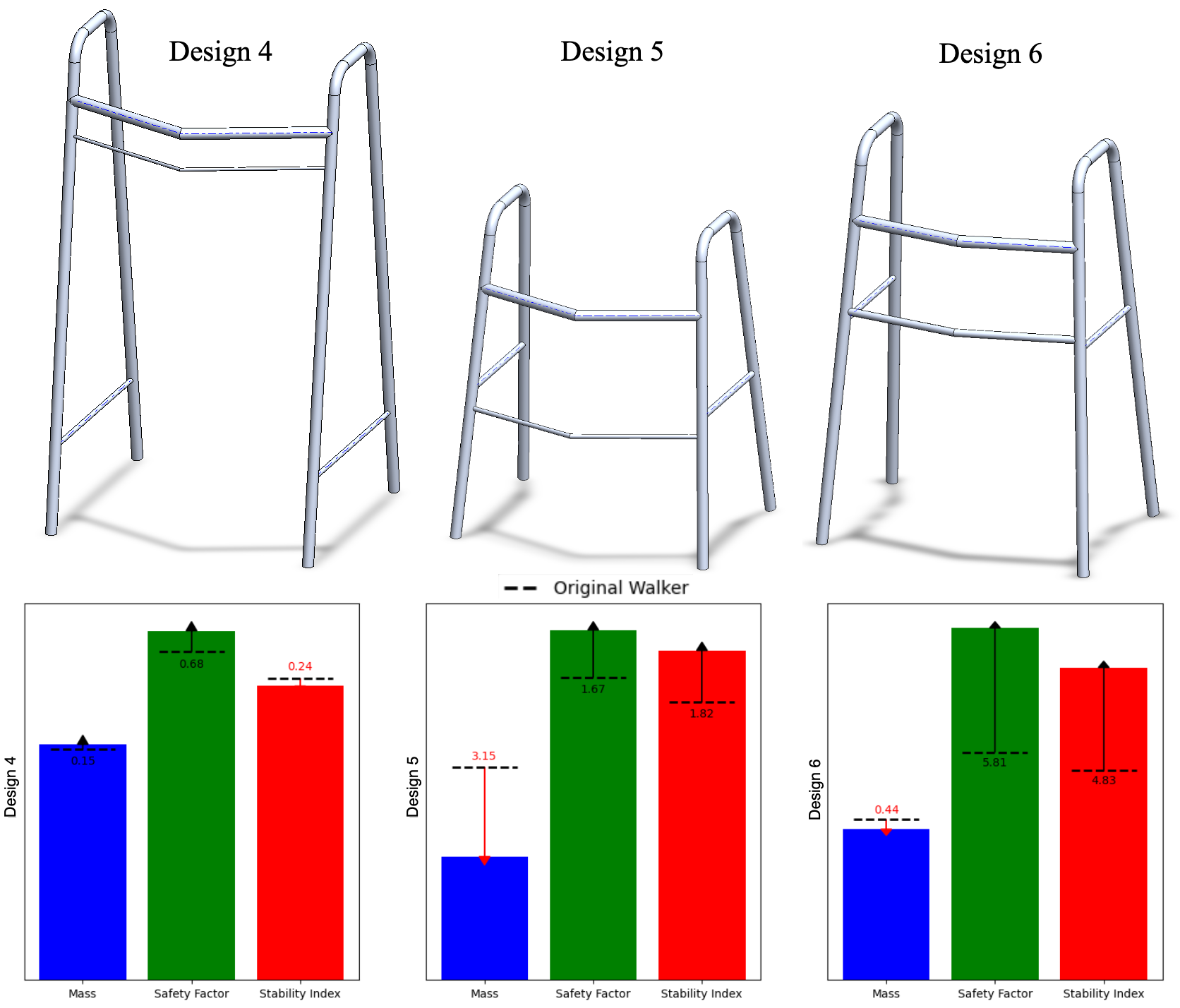}
    \caption{}
    \label{fig:customOptimizedDesigns}
\end{figure}

Please refer to Figure \ref{fig:customOptimizedDesigns} for the following case study analyses.
\\
\subsubsection*{Design 4}

This design is optimized for unusually tall people. Notably, the side crossbeams are very low compared to the rest of the walker, and the front crossbeams considerably protrude forward. Tall walkers are generally heavier and less stable. However, this design manages to balance these objectives with roughly the same weight and stability index as the original design. 
\\
\subsubsection*{Design 5}

This design can be thought of as a pediatric walker or a walker for individuals of short stature, due to its shorter height and handle distance. There is nothing too extraordinary about this design, though it demonstrates the optimizer’s capabilities of handling unique configuration optimization.
\\

\subsubsection*{Design 6}
This design was optimized for mainly stability and structural integrity. As shown in It has a much greater stability index and over 150\% more structural integrity than the original walker. We see that the overall frame resembles more of a pyramid due to its sharper angles with the vertical. However, this is expected due to the reduced mass reduction. Additionally, we observe that the lower front crossbeam is in line with the side crossbeams. This potentially contributes to the enhanced structural integrity of this walker design. This design would be suitable for overweight individuals, or individuals with balance disorders such as vertigo.

\section{Limitations/Future Work}
Our data-driven methods of optimization are the first to be applied to the optimization of a walker. However, our work contains a few key limitations discussed in this section, along with potential future work.

Our parametric model does not encompass the full spectrum of potential walker configurations. For example, the model does not support multiple crossbeams on the frame. Future works can add binary/boolean parameters corresponding to a crossbeam’s presence in a specific design, thereby increasing the expressivity of the model. As discussed in the Dataset Generation section, we acknowledge that the parametric model also does not account for the slight body rotation of the walker from the wheels, and other simplifications that may marginally stray from reality. Additionally, due to the strict computational budget, we acknowledge that the dataset may not meet industry standards. Though computationally expensive, future works may more expansively generate/simulate the dataset to improve the overall optimization accuracy. Nevertheless, all optimized designs should undergo real-world testing and clinical validation before deployment.

Regarding the optimization, to reiterate, our work considers 3 objectives: Structural Integrity, Stability, and Weight. However, we do not consider the user-centric aspects of a walker, such as the precise ergonomics and manufacturing cost. Our research can be expanded to include many more optimization objectives, increasing real-world feasibility. 

\section{Conclusion}
This study utilizes novel data-driven approaches, known for their targeted optimization capabilities, to optimize a walker, a widely used assistive device for enhanced mobility. We generate a dataset of 5,000 walkers with 8 corresponding performance values under a load case. We also validate and study the dataset using KDE plots and performance space scatterplots. Using AutoML, we train a surrogate predictive model on the dataset and achieve a relatively high average coefficient of determination of ~0.8. Additionally, we introduce a novel method of calculating the stability index of a walker. Finally, we use Multi-Objective Counterfactuals for Design, a novel genetic-based optimizer, to demonstrate its capabilities for individual custom optimization, generic optimization, and much more. We open-source all the code, models, and datasets to help researchers further develop assistive mobility devices. This work aims to demonstrate the advantages of the previously unseen applications of data-driven design on mobility devices and to take a step forward in improving and extending lives.

\section{Acknowlodgements}
I would like to acknowledge Lyle Regenwetter for his input and help throughout the project.

\newpage
\section{Appendix}
\subsection{Design Parameterization Model}
\begin{figure}[H]
    \centering
    \includegraphics[scale=0.3]{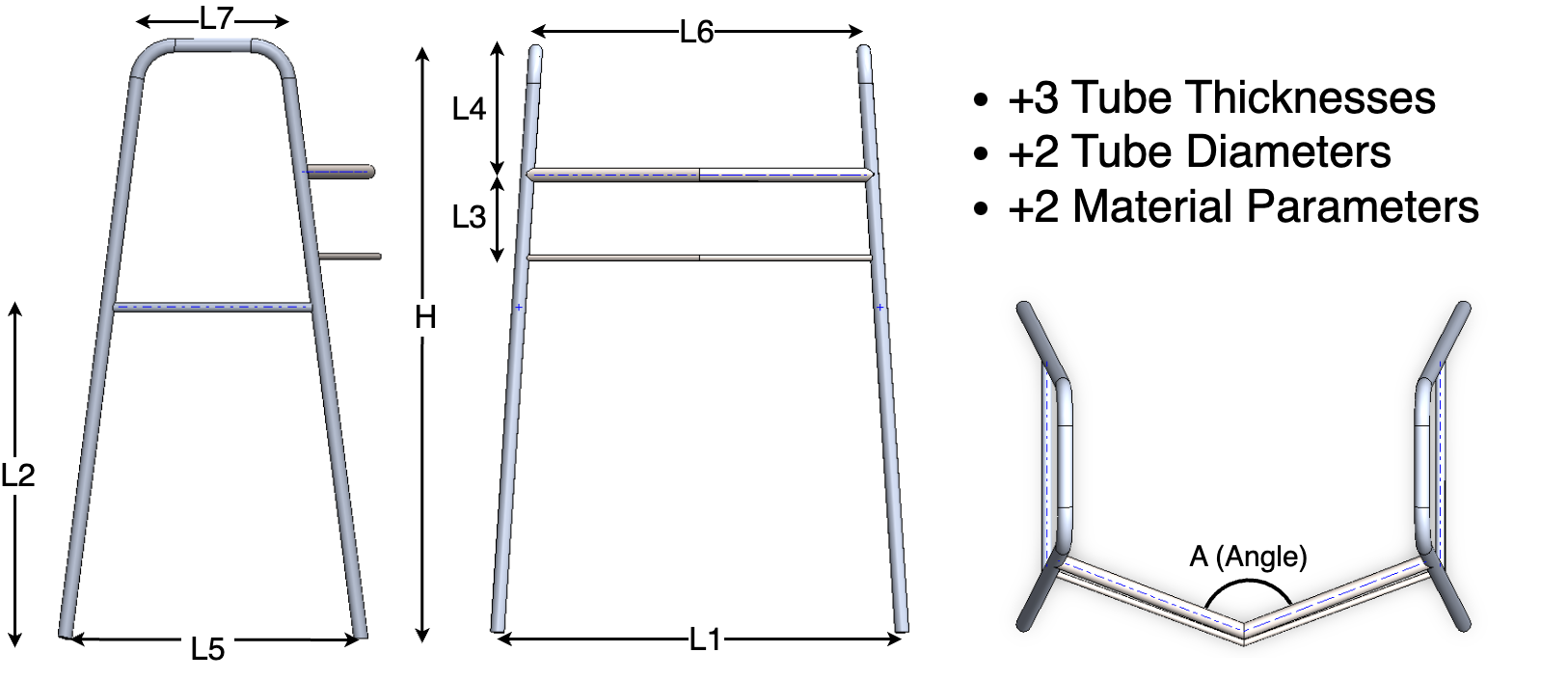}
    \caption{Design Parameters of the Walker (Excluding Tube Diameters and Material Parameters)}
    \label{fig:params}
\end{figure}

Figure \ref{fig:params} illustrates all the parameters used throughout the paper. It consists of 16 parameters, 2 of which are material parameters, and 5 of which define tube thicknesses and diameters.

\subsection{Sobol Vector Scaling}

In the sobol sequence, we initially sample 14-dimensional vectors with each continuous component ranging from 0 to 1 as the default. We then scale each vector appropriately to the given range using the following equation modified for 14 parameters:

\begin{equation*}
\begin{bmatrix}
s_{1,1}{} & \cdots  & s_{1,14}\\
\vdots  & \ddots  & \vdots \\
s_{n,1} & \cdots  & s_{n,14}
\end{bmatrix} =\begin{bmatrix}
u_{1,1}{} & \cdots  & u_{1,14}\\
\vdots  & \ddots  & \vdots \\
u_{n,1} & \cdots  & u_{n,14}
\end{bmatrix} \cdot \begin{bmatrix}
r_{1} & \cdots  & r_{14}
\end{bmatrix} +\begin{bmatrix}
l_{1} & \cdots  & l_{14}
\end{bmatrix}
\end{equation*}

where $s_{j,k}$ is the desired scaled vector of component/parameter $k$ of design $j$, $u_{j,k}$ is the unscaled vector component/parameter
$k$ of design $j$, rk is the defined range for component/parameter $k$, $l_k$ is the lower bound for component/parameter $k$, and $n$ is
the number of data points/samples.

\bibliographystyle{IEEEtran}
\bibliography{bibliography}

\begin{thebibliography}{10}
\providecommand{\url}[1]{#1}
\csname url@samestyle\endcsname
\providecommand{\newblock}{\relax}
\providecommand{\bibinfo}[2]{#2}
\providecommand{\BIBentrySTDinterwordspacing}{\spaceskip=0pt\relax}
\providecommand{\BIBentryALTinterwordstretchfactor}{4}
\providecommand{\BIBentryALTinterwordspacing}{\spaceskip=\fontdimen2\font plus
\BIBentryALTinterwordstretchfactor\fontdimen3\font minus
  \fontdimen4\font\relax}
\providecommand{\BIBforeignlanguage}[2]{{%
\expandafter\ifx\csname l@#1\endcsname\relax
\typeout{** WARNING: IEEEtran.bst: No hyphenation pattern has been}%
\typeout{** loaded for the language `#1'. Using the pattern for}%
\typeout{** the default language instead.}%
\else
\language=\csname l@#1\endcsname
\fi
#2}}
\providecommand{\BIBdecl}{\relax}
\BIBdecl

\bibitem{gell2015mobility}
N.~M. Gell, R.~B. Wallace, A.~Z. LaCroix, T.~M. Mroz, and K.~V. Patel,
  ``Mobility device use in older adults and incidence of falls and worry about
  falling: Findings from the 2011--2012 national health and aging trends
  study,'' \emph{Journal of the American Geriatrics Society}, vol.~63, no.~5,
  pp. 853--859, 2015.

\bibitem{meng2019use}
H.~Meng, L.~J. Peterson, L.~Feng, D.~Dobbs, and K.~Hyer, ``The use of mobility
  devices and personal assistance: a joint modeling approach,''
  \emph{Gerontology and Geriatric Medicine}, vol.~5, p. 2333721419885291, 2019.

\bibitem{stevens2009Unintentional}
J.~Stevens, K.~Thomas, L.~Teh, and A.~Greenspan, ``Unintentional fall injuries
  associated with walkers and canes in older adults treated in u.s. emergency
  departments,'' \emph{Journal of the American Geriatrics Society}, vol.~57,
  pp. 1464--9, 07 2009.

\bibitem{NBERw8808}
\BIBentryALTinterwordspacing
D.~E. Bloom, D.~Canning, and B.~Graham, ``Longevity and life cycle savings,''
  National Bureau of Economic Research, Working Paper 8808, February 2002.
  [Online]. Available: \url{http://www.nber.org/papers/w8808}
\BIBentrySTDinterwordspacing

\bibitem{walkerPat}
C.~William, Patent 2\,656\,874, 1949.

\bibitem{mostofa2021smart}
N.~Mostofa, C.~Feltner, K.~Fullin, J.~Guilbe, S.~Zehtabian, S.~S. Bacanl{\i},
  L.~B{\"o}l{\"o}ni, and D.~Turgut, ``A smart walker for people with both
  visual and mobility impairment,'' \emph{Sensors}, vol.~21, no.~10, p. 3488,
  2021.

\bibitem{shin2016smartwalker}
J.~Shin, A.~Rusakov, and B.~Meyer, ``Smartwalker: An intelligent robotic
  walker,'' \emph{Journal of Ambient Intelligence and Smart Environments},
  vol.~8, no.~4, pp. 383--398, 2016.

\bibitem{feurer2015efficient}
M.~Feurer, A.~Klein, K.~Eggensperger, J.~Springenberg, M.~Blum, and F.~Hutter,
  ``Efficient and robust automated machine learning,'' \emph{Advances in neural
  information processing systems}, vol.~28, 2015.

\bibitem{regenwetter2023framed}
L.~Regenwetter, C.~Weaver, and F.~Ahmed, ``Framed: An automl approach for
  structural performance prediction of bicycle frames,'' \emph{Computer-Aided
  Design}, vol. 156, p. 103446, 2023.

\bibitem{regenwetter2023counterfactuals}
L.~Regenwetter, Y.~A. Obaideh, and F.~Ahmed, ``Counterfactuals for design: A
  model-agnostic method for design recommendations,'' \emph{arXiv preprint
  arXiv:2305.11308}, 2023.

\end{thebibliography}

\end{document}